\documentclass[10pt,twocolumn,letterpaper]{article}

\usepackage{iccv}
\usepackage{times}
\usepackage{epsfig}
\usepackage{graphicx}
\usepackage{amsmath}
\usepackage{amssymb}
\usepackage{algorithmicx,algorithm}
\usepackage{multirow}
\usepackage{subfigure}
\usepackage{bm}
\usepackage{ctable}
\usepackage{lipsum}

\usepackage[pagebackref=true,breaklinks=true,letterpaper=true,colorlinks,bookmarks=false]{hyperref}

\iccvfinalcopy 

\ificcvfinal\fi

\begin{document}

\title{H2O: A Benchmark for Visual Human-human Object Handover Analysis}

\author{Ruolin Ye*
\and Wenqiang Xu*
\and Zhendong Xue
\and Tutian Tang
\and Yanfeng Wang 
\and Cewu Lu\\
Shanghai Jiao Tong University\\
{\tt\small \{cathyye2000, vinjohn, 707346129, tttang, wangyanfeng, lucewu\}@sjtu.edu.cn}
}

\maketitle
\newcommand\blfootnote[1]{%
\begingroup
\renewcommand\thefootnote{}\footnote{#1}%
\addtocounter{footnote}{-1}%
\endgroup
}
\blfootnote{* These two authors have equal contribution.}
\ificcvfinal\thispagestyle{empty}\fi

\begin{abstract}
  Object handover is a common human collaboration behavior that attracts attention from researchers in Robotics and Cognitive Science. Though visual perception plays an important role in the object handover task, the whole handover process has been specifically explored. In this work, we propose a novel rich-annotated dataset, H2O, for visual analysis of human-human object handovers. The H2O, which contains 18K video clips involving 15 people who hand over 30 objects to each other, is a multi-purpose benchmark. It can support several vision-based tasks, from which, we specifically provide a baseline method, RGPNet, for a less-explored task named Receiver Grasp Prediction. Extensive experiments show that the RGPNet can produce plausible grasps based on the giver's hand-object states in the pre-handover phase. Besides, we also report the hand and object pose errors with existing baselines and show that the dataset can serve as the video demonstrations for robot imitation learning on the handover task. Dataset, model and code will be made public.
\end{abstract}

\section{Introduction}
\begin{figure}[t!]
\centering
\includegraphics[width=0.9\linewidth]{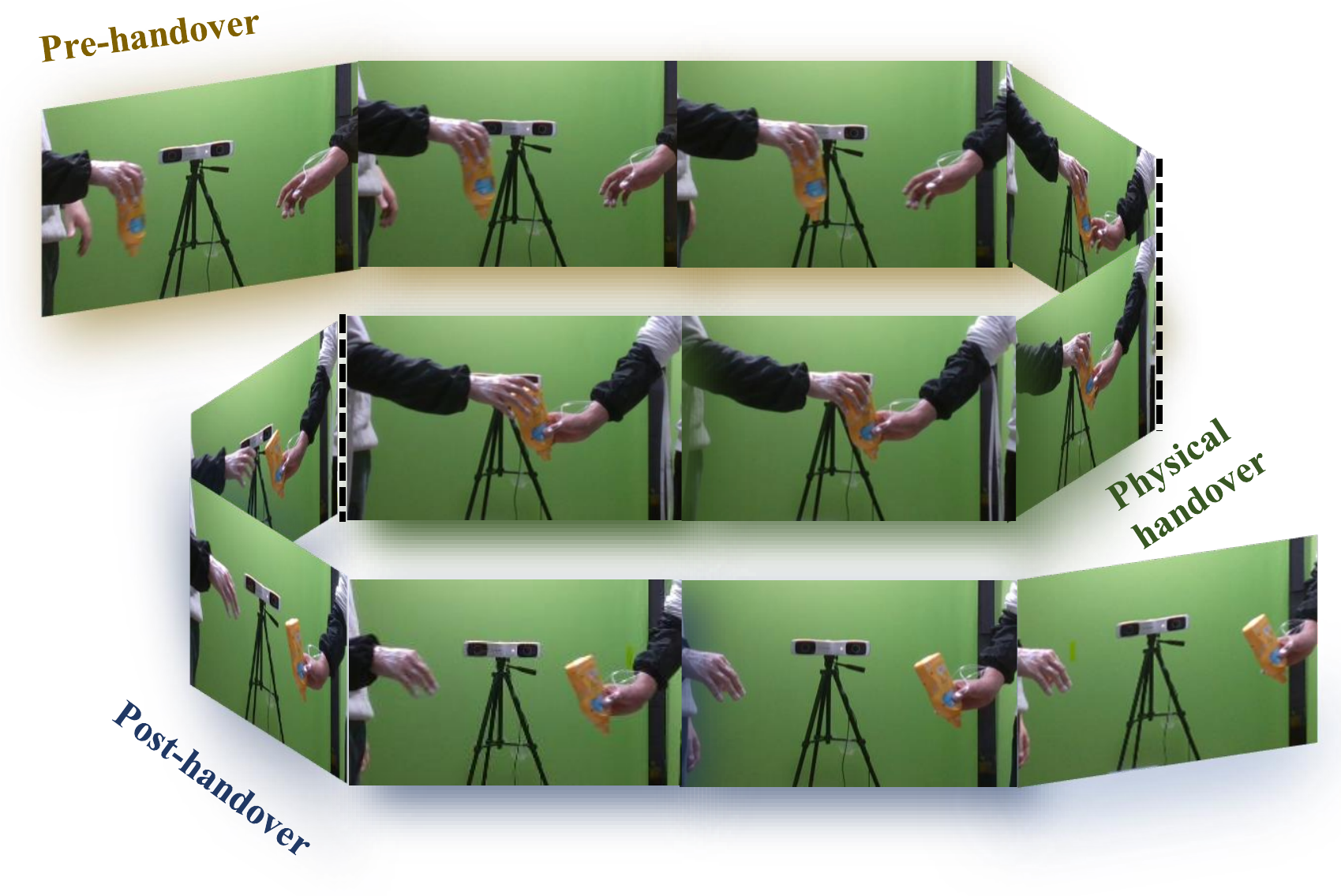}

   \caption{A process of handover typically includes three stages, namely pre-handover phase, physical exchange phase, and post-handover phase.}
\label{fig:intro}
\end{figure}

Object handover is one of the basic human collaboration behaviours yet a complex process in which the giver transfers an desired object to the receiver (Fig. \ref{fig:intro}). It is not only a rich area for the robotics community \cite{handover_robot_human_1} and the cognitive community \cite{handover_cog1}, but also for the computer vision community. Various perception behaviours are involved in this process, including but not limited to object detection (\textit{where to find the object}) \cite{faster_rcnn, mask_rcnn}, visual grasp prediction (\textit{how to grasp the object}) \cite{ganhand}, pre-handover reconstruction (\textit{how the giver interact with the object during passing over}) \cite{ho_recon_photo, ho_recon1} and receiver grasp prediction (\textit{how the receiver obtain the object}).

Though exploring the object handover task can be beneficial for all these three communities mentioned above, it has seldom been explored from a pure computer vision perspective.
The main reason might be the lack of specialized datasets. The currently available handover datasets are either of small scale \cite{handover_data1} or have no visual recordings \cite{handover_data2}. On the other hand, the researchers from the computer vision community have just recently paid their attention to the single hand-object interaction reconstruction task \cite{obman, ho_recon_photo, ho_recon1}. Visual analyses of how two hands interacting with objects are in the line of such works, but of higher challenges and practical impact.

\begin{figure}[t!]
\centering
\includegraphics[width=0.95\linewidth]{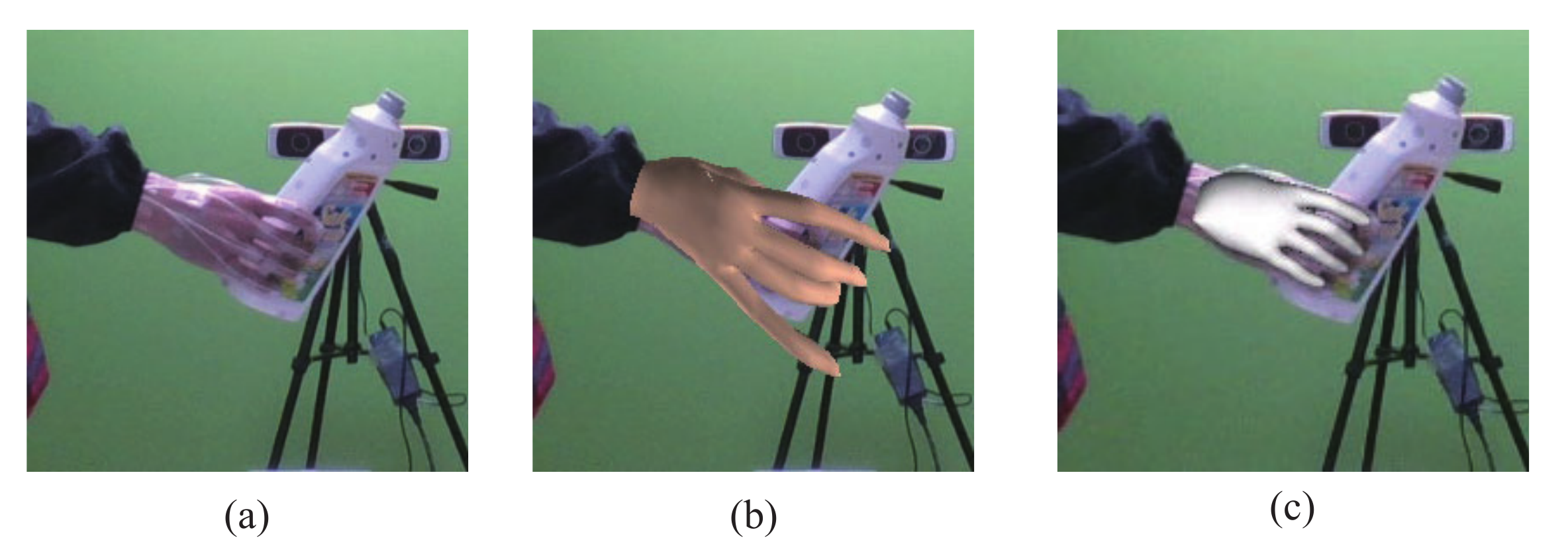}
   \caption{Image content: (a) original RGB image; (b) hand model recovered from markerless setting via a similar approach described in \cite{ho3d}; (c) hand model recovered from marker-based recording. As we can see from the above images, markerless fitting is easily unstable even when the grasp type is fairly simple, which can be catastrophic for trajectory recording.}
\label{fig:marker_reason}
\end{figure}

In this work, we propose a rich-annotated knowledge base about the \textbf{H}uman-\textbf{H}uman \textbf{O}bject handover task, named \textbf{H2O}. To construct the H2O dataset, we invite 15 volunteers to form 40 giver-receiver pairs. Each pair pass over 30 objects and this results in a total of 18K videos clips and 5M frames with rich annotations, such as hand pose, object pose, handover stage, grasp type, task-oriented intention, \etc. Since the handover process involves many complex or dynamic grasps which are hard to be regressed by the current state-of-the-art hand model reconstruction method (See Fig. \ref{fig:marker_reason}), we determine to adopt a marker-based protocol to record both the dynamics of hand and object poses. Being aware of the markerless preference for some research topics \cite{ho3d, contactpose}, we also provide a photorealistic H2O-Syn dataset, where the poses of hand and object are transferred from H2O, and the textures of hand and object are scanned from the real counterparts. As human beings are generally the experts that an intelligent robot should learn the skills from, we believe this human-human interaction dataset can also serve as the demonstrations of various manipulation tasks for robots, such as human-robot handover \cite{handover_human_robot1, handover_human_robot2}, robot-human handover \cite{handover_robot_human_1, handover_robot_human_2, handover_robot_human3}, and dexterous manipulation \cite{dexterous_manpulation_il1, dexterous_manpulation_il2}.  

As mentioned earlier, many handover-related vision tasks have been explored by previous researchers, while the \textit{Receiver Grasp Prediction} (\textbf{RGP}) task is seldom explored, mostly because it requires a dataset that contains at least two hands and one object interacting with each other. Thus, in this paper, we take a look into this problem to enrich the visual analysis of the whole handover process.  
To address the RGP problem, the model should be able to 1). estimate the hand-object interaction model from a single RGB image during the pre-handover phase; 2). reason the available regions to predict a plausible grasp configuration; 3). refine the receiver hand configuration to avoid interpenetration or hindering the retract path of the giver's hand. To accomplish that, we propose a generative approach named \textbf{RGPNet} to predict plausible receiver grasps. The RGPNet will first predict the hand and object pose from the pre-handover image. Then, the pose information along with the contextual information will be encoded as the condition vector, which will then be forwarded to predict the relative object and hand pose from the predicted object and giver's hand pose. Predicting the relative pose, instead of the absolute one, can alleviate the collision between the two hands (Sec. \ref{sec:rgp}).

To demonstrate that our proposed H2O can help benchmark multiple handover-related tasks, we not only report the grasp score \cite{grasp_score} and interpolation volume \cite{obman} for the RGP task, but also report the hand and object pose error for the hand-object interaction reconstruction in the pre-handover subset of the H2O. Besides, we show the learned RGPNet has the generalization ability to the ContactPose sub-dataset which has hand-off annotation. Finally, we showcase the ability of our dataset to be the video demonstrations for imitation learning in simulation.

We summarize our contributions as follows:
\begin{itemize}
    \item We propose a novel dataset that can support the comprehensive visual analysis of the object handover process. It can help mitigate the gap between current vision research and robotics researches. Among all supported tasks, we report baseline performance on the hand-object interaction reconstruction task and the receiver grasp prediction task. Besides, we illustrate that H2O recordings can be used as video demonstrations for robot learning.
    \item We propose a novel generative pipeline RGPNet to address the challenging receiver grasp prediction (RGP) task. The RGPNet learned from the H2O dataset can be generalized to the ContactPose hand-off subset.
\end{itemize}

\section{Related Work}
\paragraph{Datasets for Handover} To the best of our knowledge, currently there exist three public datasets \cite{handover_data1, handover_data2, handover_sr} available for the handover tasks. Our proposed H2O is closely related to \cite{handover_data1} for it is the only one who has the visual observations. However, it contains only 1110 video clips but no vision-related annotations, which limits the applicability for visual analysis. \cite{handover_data2} directly learns the human behaviour from the 3D skeleton trajectories recorded by MoCap devices. Besides, they show more interest in the whole body movement but not the hand, which limits the diversity on hand-object manipulation. \cite{handover_sr} collects 5202 grasps to analyze the grasp type choice when human handover objects to each other. However, it does not provide any visual observations.

The datasets which are built for hand-object interaction modeling \cite{obman, fphab, ho3d, contactpose} are related but not sufficient. Since most of them involve only one hand and not handover-intended, such datasets might serve as valuable pretraining materials for visual analysis of pre-handover and post-handover phase, but not quite adequate for the physical handover phase.

More comprehensive comparison among datasets can be found in Table \ref{tab:dataset_comp}.

\paragraph{Perception in Handover task}
A handover task is a complex task, which can be addressed from several different perspectives of approaches, including perception, planning, and control. In this paper, we are particularly interested in the perception problems related to the handover task. The perception in handover tasks involves first-person perspective and third-person perspective.

\begin{figure*}[]
\centering
\includegraphics[width=\linewidth]{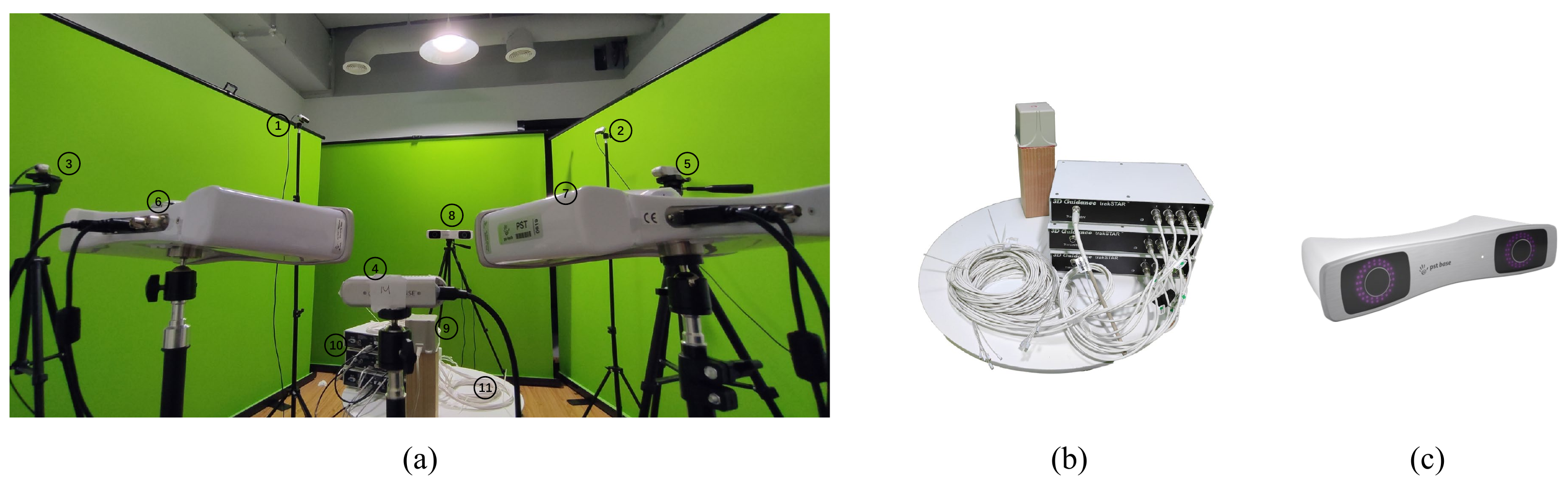}

   \caption{The recording setup: (a). The whole recording scene: Devices 1 and 2 are the first-person perspective cameras; 3-5 are the third-person perspective cameras; 6-8 are the PST base object tracking devices; 9-11 are the NDI trackSTAR hand pose tracking devices. To be specific, 9 is the transmitter, 10 are the electronics units, 11 are the sensors; (b). Hand pose tracker; (c). Object pose tracker}
\label{fig:recording_setup}
\end{figure*}
For the \textbf{first-person perspective}, we will elaborate on the giver's viewpoint and receiver's viewpoint respectively.

During the pre-handover phase, the giver should try to locate the object requested by the receiver, which corresponds to object detection \cite{faster_rcnn, mask_rcnn}, then the giver should try to grasp it into the hand, which can be handled by grasp prediction \cite{ganhand}. After that, the giver should determine where the receiver will likely receive the object, the visual task of this procedure is human/hand detection or tracking \cite{human_det_track}. 

During the physical exchange phase, the receiver will determine how to obtain the object (\textit{receiver grasp prediction}) based on the information such as object pose, affordance, giver's grasp or the contextual content. After the giver release the object, the receiver will turn to use the object for the task he was thinking.

For the \textbf{third-person perspective}, the visual system which analyzes the human-human handover process can be used for demonstration of robot-related tasks, such as human-robot handover \cite{handover_human_robot1, handover_human_robot2}, robot-human handover \cite{handover_robot_human_1, handover_robot_human_2, handover_robot_human3}, dexterous manipulation \cite{dexterous_manpulation_il1, dexterous_manpulation_il2}, \etc. Besides, it can also serve as a analysis tool for cognitive study on human behaviour understanding \cite{handover_cog1, handover_sr}.

\section{The H2O Handover Dataset}
In this section, we will describe the protocol of the data collection process.

\subsection{Acquisition of Real Data}

\begin{figure}[]
\centering
\includegraphics[width=0.95\linewidth]{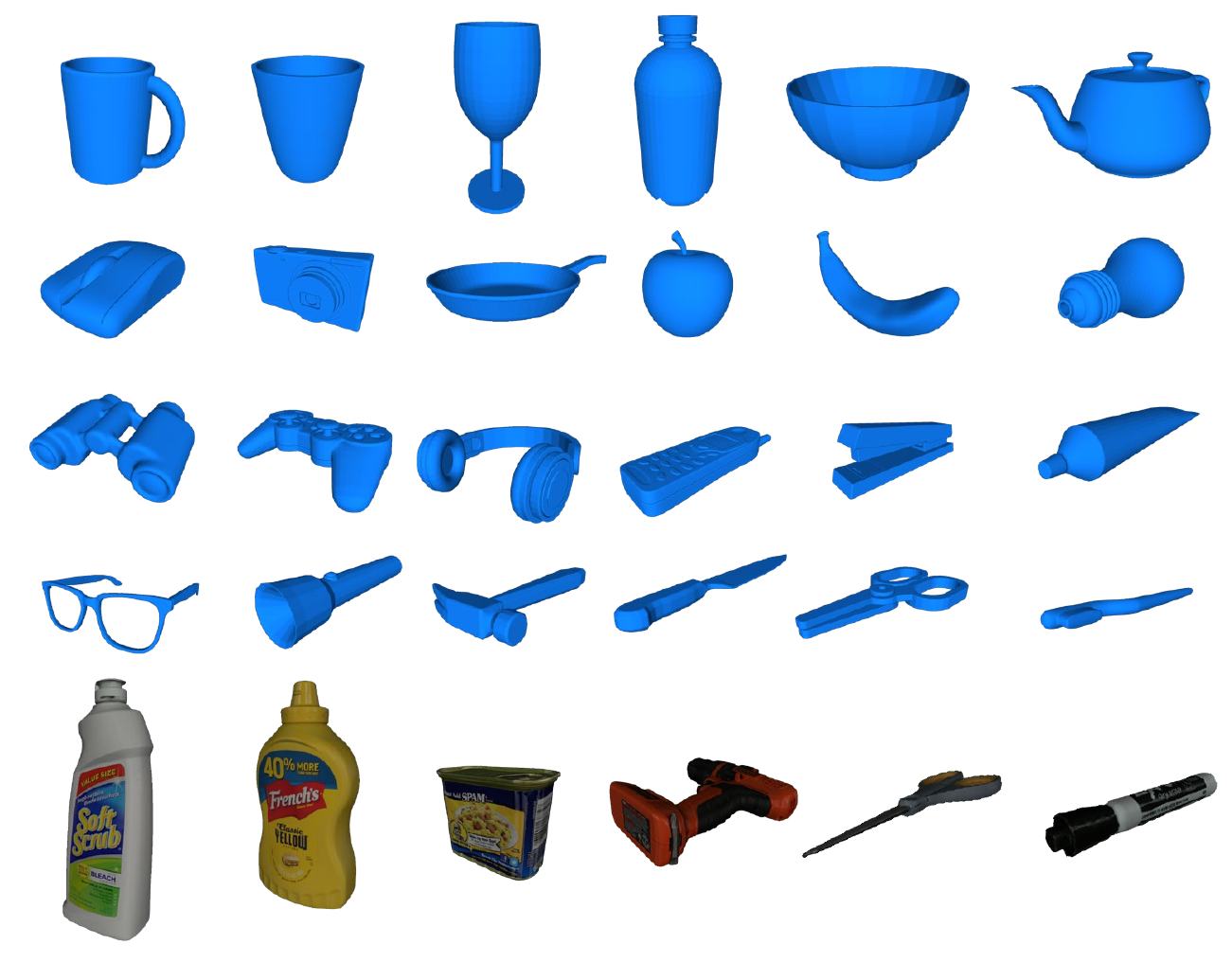}

   \caption{Object models from ContactPose \cite{contactpose} (upper), object model from YCB object \cite{ycb} (bottom).}
\label{fig:obj_repository}
\end{figure}

\paragraph{Object Repository}
We adopt all the object models used in ContactPose paper \cite{contactpose} and select some of the YCB objects. Upon submission, we have a total of 30 objects in the repository to be passed over, which are displayed in Fig. \ref{fig:obj_repository}. The choice of the object models is based on the fact that existing datasets like \cite{contactpose, ganhand} have already annotated much useful information on these models. Though our dataset is self-contained, these datasets can be used as a great source for additional data augmentation. 

\paragraph{Devices and Setup} The whole recording setup is demonstrated in Fig. \ref{fig:recording_setup}. We record the RGB-D videos with 5 Intel Realsense D435 cameras\footnote{https://www.intelrealsense.com/depth-camera-d435/}. Among them, 3 are from third-person perspective (denoted as ``TPP") and the rest 2 are from first-person perspective (denoted as ``FPP"). The resolution is $1280\times 720$ at $30$ fps.

We adopt 12 NDI TrakSTAR magnetic sensors\footnote{https://www.ndigital.com/products/3d-guidance/} to record the hand poses, among which 5 are for each hand tip and 1 for each dorsum. The recording frequency is $60$ fps. The sensor is 8 mm wide.

The object poses are recorded by 3 PST Base optical trackers\footnote{https://www.ps-tech.com/products-pst-base/}, which are in triangular arrangement and synchronized to each other, so that they can capture the markers from multiple viewpoints. The recording frequency is $30$ fps. Every object should be registered by the tracking system so that the pose can be properly recorded.

The scene is surrounded by three green screens, which makes it easier for others to change the background whenever necessary.

\begin{figure*}[]
\centering
\includegraphics[width=0.95\linewidth]{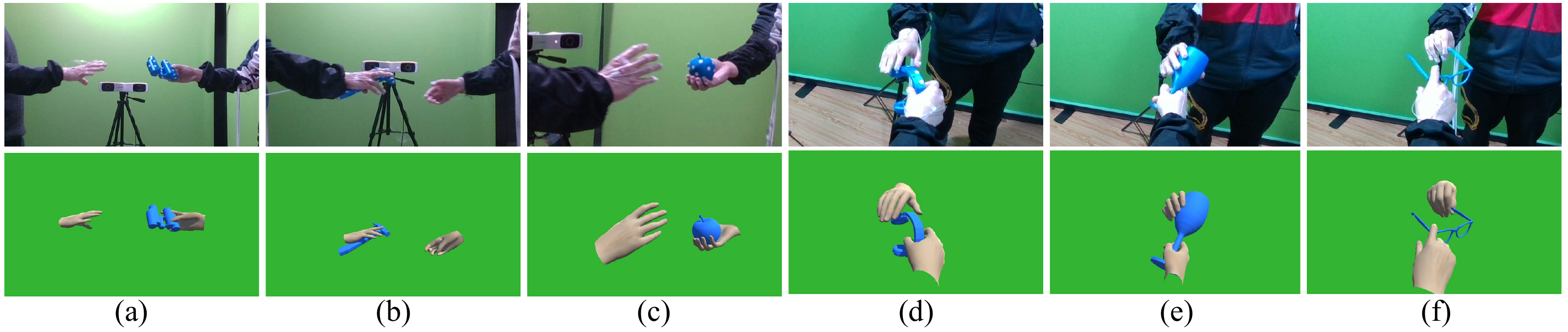}

   \caption{Samples from H2O Dataset (Upper) and H2O-Syn Dataset (Bottom): (a) (b) are samples from the middle TPP camera; (c) is from the left TPP camera; (d) (e) (f) are samples from FPP cameras.}
\label{fig:dataset_sample}
\end{figure*}

\paragraph{Recording Setup}A director will help attach the magnetic sensors to the volunteers' hands and give instructions during the recording process. For each pair of volunteers, the receiver should ask the giver to pass over a specific object. The receiver should be made clear the usage of the requested objects. The director will also supervise the process and monitor the screen so that the handover trajectories are in a proper position.

\subsection{Annotation Protocol}

\paragraph{Hand Pose} The hand pose encodes 21 DoF of hand configuration, which is recorded similarly as the procedures described in \cite{bighand}. We attach 6 magnetic trackers to each finger and the dorsum on one hand. And the rest 15 joint location is calculated from hand priors and inverse kinematics. The recorded hand pose is projected to image camera space by an iterative PnP algorithm implemented in OpenCV \cite{opencv}. To note, all the volunteers are right-handed.

\paragraph{Object Pose} The object 6DoF pose is tracked by the retro-reflective markers attached to the object body. The recording procedure follows the instruction from the user manual. To note, though we adopt a multi-view setting, the markers may still be missing due to severe occlusion during handover. For this situation, we manually annotate the 6D pose.

\paragraph{Handover Stage} For each video clip, we separate it into three different phases, namely the pre-handover, physical handover, and post-handover. In the pre-handover process, the giver grasps the object and tries to pass it over to the receiver. While the physical handover phase starts when the receiver attaches the object and ends until the giver retracts from the object. After that, it moves on to the post-handover phase, during which each participant will continue their task before the handover. 

\paragraph{Grasp Type} We annotate each frame according to the grasp taxonomy described in \cite{grasp_taxonomy}. We find such annotations could be useful for human behaviour understanding like in \cite{handover_sr, fphab}, and grasp prediction like in \cite{ganhand}. 

\paragraph{Task-Oriented Intention} We also take notes on whether this grasp is task-oriented. During recording, before pre-handover, the receiver will tell the giver how he/she would use the object for. If the receiver mentions "casual", then the giver will pass the object however he/she prefers. For the task list concerning each object, please refer to the supplementary materials.

\subsection{H2O-Syn Dataset}
For those researchers who prefer marker-less RGB datasets, we adopt the Unity engine\footnote{https://unity.com/} to create a H2O-Syn dataset. Though the dataset is created from the CG renderer, we try to make it photorealistic by: (1). Adopting the physical camera in Unity, so that once the object and hand poses are properly transferred and placed in the Unity editor, the depth image from the real world can be directly aligned; (2). The textures of hands and objects are all scanned from the real world. Some samples of H2O-Syn datasets are displayed in Fig. \ref{fig:dataset_sample}.

\subsection{Statistics}\label{sec:statistic}
In this section, we first report the basic statistics of the H2O dataset and demonstrate how it is prepared for the hand-object interaction reconstruction task and RGP task. Finally, to show the diversity of the dataset, we show figure results on the hand and object pose distributions.

\paragraph{Train/Val/Test Split For Receiver Grasp Prediction}
Since the frames are highly correlated within a video clip, we split the dataset according to handover behaviors. Among all of the 18K video clips corresponding to 6K handovers, we randomly select 5K for training, 500 for validation, and 500 for testing. For the pre-handover subset, we have a total of 2.5M images. When training the model for hand-object pose estimation and receiver grasp prediction, we sample the images one every 10 frames, so we have 210K for training, 21K for validation, and 21K for testing. For the physical handover subset, we have a total of 1M images. The main reason for the decrease in the number of frames is because the physical exchange phase usually finishes quicker than the pre-handover. From the object-centric perspective, we have an average of 7K frames for each object to train in the pre-handover phases.

For a more detailed description of image split and statistics, please refer to the supplementary files.

Since the H2O-Syn dataset is one-to-one transferred from the H2O dataset, the split strategy is the same for the H2O-Syn dataset. 

\begin{figure}[]
\centering
\includegraphics[width=\linewidth]{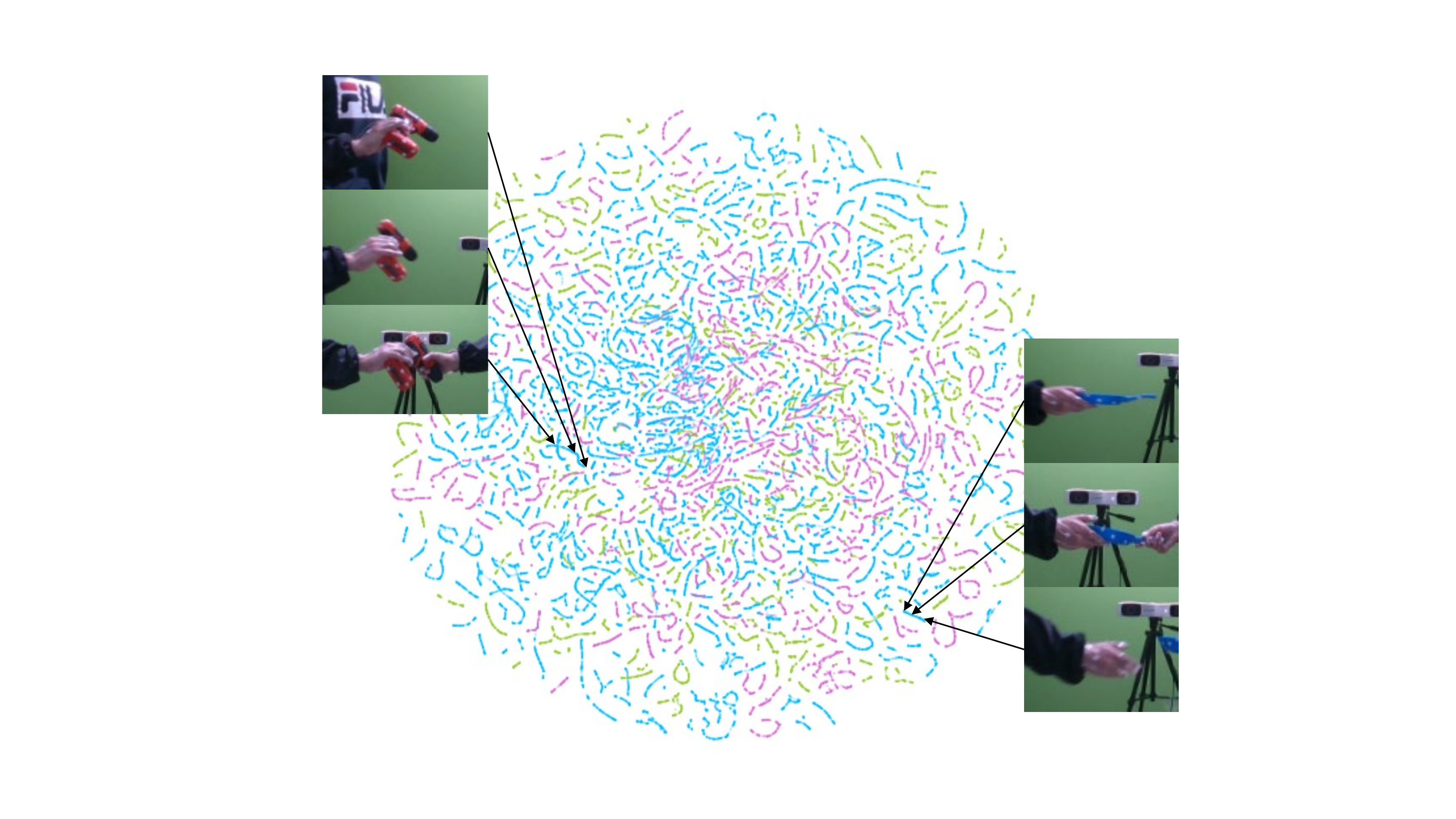}

   \caption{Hand pose distribution: Visualization of t-SNE embedding of hand poses. A point stands for a data sample, and a sequence with the same color represents a handover motion sequence. Typical motion sequence of power drill and pan are shown in the figure.}
\label{fig:hand_pose_dist}
\end{figure}

\paragraph{Hand Pose Distribution}
We project the hand pose distribution from the H2O dataset with t-SNE \cite{tsne}. As we can see from Fig. \ref{fig:hand_pose_dist}, the hand poses from the same video clip are close to each other and form a segment on the 2D projected image, while the hand pose of large configuration variations is far apart.

\begin{table*}[]
\centering
\begin{tabular}{ccccccccccc}
\hline
\small{Name}            & \begin{tabular}[c]{@{}c@{}}\small{\# Sub-}\\\small{jects}\end{tabular} & \begin{tabular}[c]{@{}c@{}}\small{\# Objects}\\\small{with Pose}\end{tabular} & \small{\# Frame }& \small{Real} & \begin{tabular}[c]{@{}c@{}}\small{Multiple}\\\small{ View-}\\\small{points}\end{tabular} & \begin{tabular}[c]{@{}c@{}}\small{First}\\\small{ Person }\\\small{View}\end{tabular} & \begin{tabular}[c]{@{}c@{}}\small{Task}\\\small{ Oriented}\end{tabular} & \begin{tabular}[c]{@{}c@{}}\small{Grasp}\\\small{ Type}\end{tabular} & \begin{tabular}[c]{@{}c@{}}\small{Handover}\\\small{ Stage}\end{tabular}  \\ \hline
\small{\textit{Single hand-object track}} \\
YCB Affordance  & -           & 21                   & 0.133M  & \checkmark    & -                    & -                 & -             & \checkmark          & -              \\
FPHA           & 6           & 4                    & 0.100M  & \checkmark    & -                    & \checkmark                 & \textbf{-}    & \checkmark          & -              \\
HO3D           & 10          & 10                   & 0.078M   & \checkmark    & \checkmark                    & -                 & -             & -          & -              \\
ObMan          & 20          & 2.7k                 & 0.15M  & -    & \checkmark                    & -                 & -             & -          & -              \\
ContactPose*    & 50          & 25                   & 2.9M  & \checkmark    & \checkmark                    & -                 & \checkmark             & -          & -              \\
H2O P-H (Ours) & 15          & 30                   & 2.5M    & \checkmark    & \checkmark                    & \checkmark                 & \checkmark             & \checkmark          & \checkmark \\\hline
\small{\textit{Handover track}} \\
MSD & 18 & - & 0.133M & \checkmark & - & - & - & - &-\\
H2O (Ours)  & 15          & 30                   & 5M    & \checkmark    & \checkmark                    & \checkmark                 & \checkmark             & \checkmark          & \checkmark              \\ \hline
\end{tabular}
\caption{Comparison with different datasets on the number of volunteers, number of objects with pose annotations, number of video frames, whether the dataset is built from real, whether have multi-view images, whether have the first-person view in images, whether the receiver's task-oriented intention is recorded, whether the grasp type is annotated, and whether the handover stages are split. \textbf{H2O P-H} means the pre-handover subset of the H2O dataset. Our proposed dataset has richer annotations and more images, which can support various tasks.  *: ContactPose dataset also has two-hand-object annotations, however, these annotations are for bimanual manipulations, thus we also count the bimanual annotation as a single hand-object track.}\label{tab:dataset_comp}
\end{table*}

\begin{figure}[]
\centering
\includegraphics[width=0.85\linewidth]{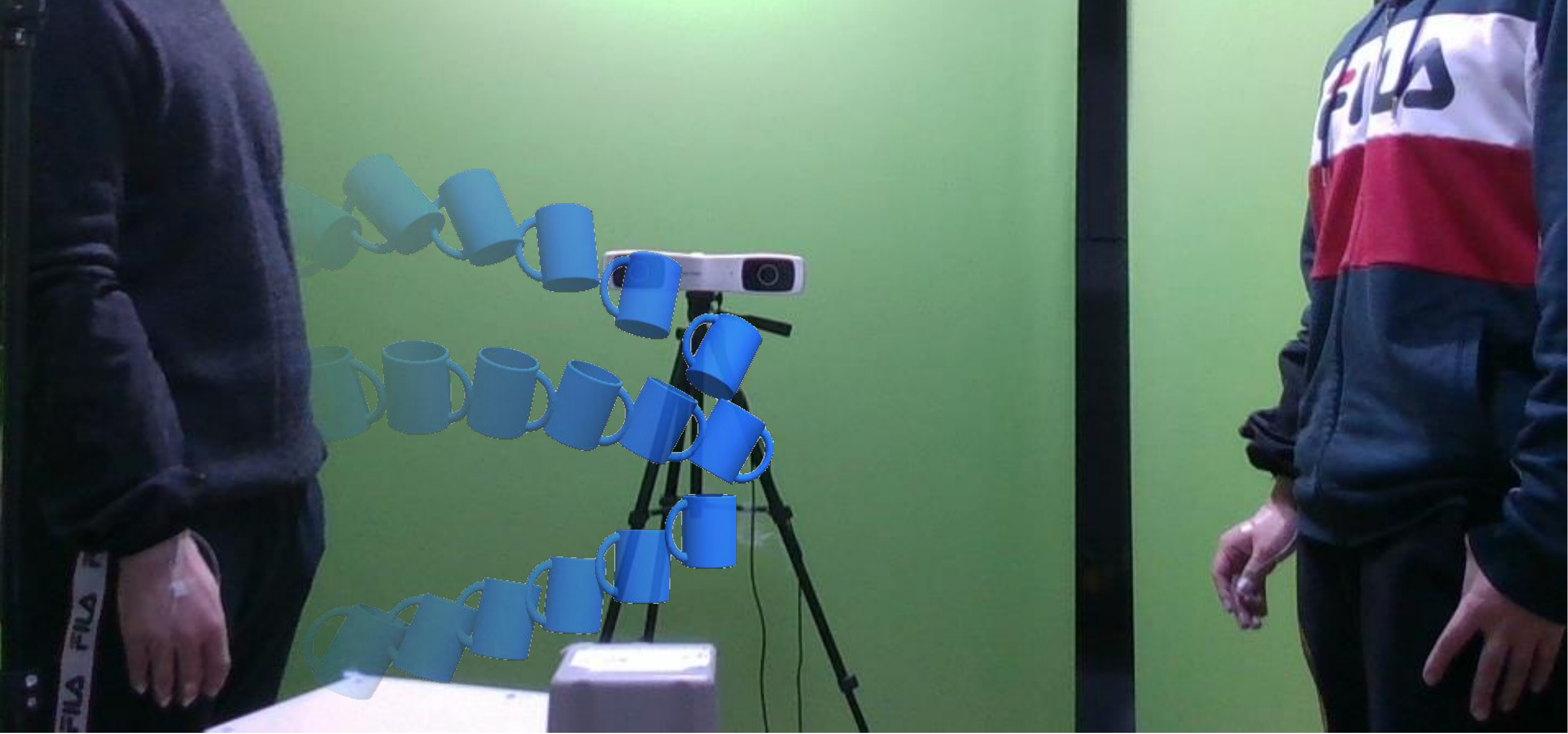}

   \caption{Object pose distribution: Visualization of three typical object pose trajectories in the pre-handover phase. The object pose varies in different trajectories in position and rotation.}
\label{fig:obj_pose_dist}
\end{figure}

\paragraph{Object Pose Distribution}
To show the diversity of object poses, in Fig. \ref{fig:obj_pose_dist}, we display the different trajectories of the same object from the giver to the receiver when different task-oriented intentions are considered. For each trajectory, even the hand-object are nearly relatively static to each other, the object poses still show fairly large variations. 

\paragraph{Comparison With Related Datasets}
We compare our H2O dataset with related datasets, which are divided in two tracks, namely single hand-object track (for its applicability to pre-handover phase) and handover track. As shown in Table \ref{tab:dataset_comp}, for the single hand-object track, FPHAB \cite{fphab}, HO3D \cite{ho3d}, ObMan \cite{obman}, ContactPose \cite{contactpose} and YCB Affordance \cite{ganhand} are compared. For the handover track, the multi-sensor dataset (MSD) \cite{handover_data1} are compared.

\begin{figure*}[]
\centering
\includegraphics[width=0.95\linewidth]{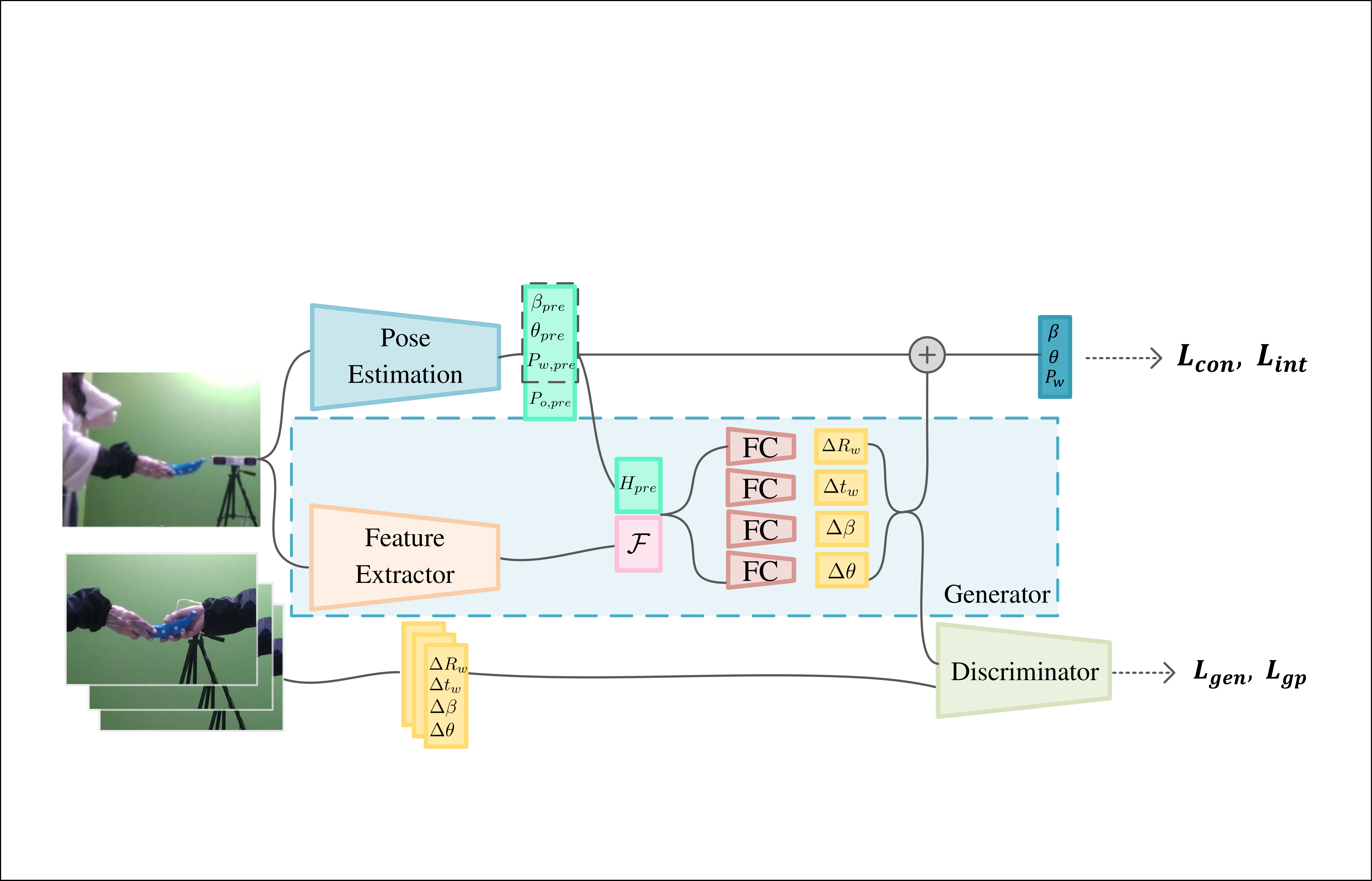}
   \caption{The overall pipeline of receiver grasp prediction.}
\label{fig:pipeline}
\end{figure*}

\section{RGPNet}
In this section, we describe the RGPNet for the receiver grasp prediction task. To note, for this task, we should first transform the images from the pre-handover phase into a hand-object predominant patch. For detailed instructions to extract the hand-object area from the full image, please refer to the supplementary materials. After that, the system will reconstruct the hand-object interaction models from the image patch $\mathcal{I}$ (Sec. \ref{sec:hoi_recon}), predict the grasp for the receiver in a generative manner, and refine it according to the constraints (Sec. \ref{sec:rgp}). The overall pipeline is illustrated in Fig. \ref{fig:pipeline}.

\subsection{Hand-object Interaction Reconstruction}\label{sec:hoi_recon}
Given an image patch $\mathcal{I}$ and the object type $C$ from the pre-handover phase, a joint hand-object pose estimation framework is adopted to predict the hand pose $H_{pre}=\{\beta_{pre}, \theta_{pre}, P_{w, pre}\}$ and object 6D pose $P_{o, pre}=\{R_{o, pre},t_{o,pre}\}$, where $\beta_{pre}$ and $\theta_{pre}$ follow the convention of MANO \cite{mano}, $P_{w, pre}=\{R_{w,pre}, t_{w, pre}\}$ denotes the wrist pose and $R_{o, pre},R_{w, pre}\in SO(3), t_{o,pre}, t_{w, pre}\in\mathbb{R}^3$. For the hand-object pose estimation framework, we provide two baselines \cite{ho_recon1, ho_recon_photo} in Table \ref{tab:hor}. For later grasp prediction experiment, we adopt the \cite{ho_recon_photo} because it has better performance on our dataset.

To obtain the 3D hand model $\mathcal{M}_{H,pre}$ and the object hand model $\mathcal{M}_{O, pre}$, the hand parameters will forward through the MANO layer \cite{obman} and the object model will be retrieved and transformed according to the object category $C$ and the predicted $P_{o,pre}$.

\subsection{Receiver Grasp Prediction, RGP-GAN}\label{sec:rgp} 
We consider the receiver's grasp as a relative transformation from the $\mathcal{M}_{H, pre}$. Thus, the network is required to predict the 6DoF transformation of the hand root $\Delta P_w=\{\Delta R_w, \Delta t_w\}$, and the shape parameter deformation $\Delta \beta, \Delta \theta$. Since there are no ground truths of the receiver's grasp for the pre-handover images, we propose to learn the parameters in a generative manner.

We first collect a set of samples from the corresponding physical handover phases of the pre-handover image, and calculate $S=\{(\Delta P_{w,i}, \Delta \beta_i, \Delta \theta_i)\mid i=1,\ldots,N\}$, where $N$ is the sample number. The $\beta$ and $\theta$ for the two hands in the physical handover images are calculated by a pre-trained SIKNet as described in \cite{bihand}.

Then, for each $\mathcal{I}$, a ResNet-18 \cite{resnet} backbone is adopted to transform it into a global feature $\mathcal{F}$. Later, $\mathcal{F}$ will be concatenated with $H_{pre}$ and input to four separate branches to predict the $\Delta R_w, \Delta t_w, \Delta \beta, and \Delta \theta$ respectively. Each branch is constituted by a 3-layer fully connected network with ReLU activation.

\paragraph{Grasp Generation Loss} With $S$, we learn the grasp transformation with a discriminator $D$ to judge if the transformation is realistic. Wasserstein loss \cite{wgan} is used. The discriminator is a 3-layer fully connected network with ReLU activation. We denote the ResNet-18 and the following fully connected layers as G (as shown in Fig. \ref{fig:pipeline}). The loss is formulated as:
\begin{equation}
    \mathcal{L}_{gen} = -\mathbb{E}_{S \sim p(S)}[D(G(\mathcal{I}))] + \mathbb{E}_{S\sim p(S)}[D(S_i)].
\end{equation}
We also adopt the gradient penalty loss $\mathcal{L}_{gp}$ \cite{wgan_gp} to guarantee the satisfaction of the Lipschitz constraint in the W-GAN \cite{wgan}.  

\paragraph{Contact Loss}
As the receiver should make contact with the object surface but not the giver's hand, we adopt the concept of anchor point from \cite{cpf}. We define a set of anchor points $\mathcal{A}^{(g)}=\{A_i^{(g)}\}_{i=1}^{17}$ of giver's hand mesh model $\mathcal{M}^{(g)}=\{v_j^{(g)}\}_{j=1}^N$, where $A_i, v_j^{(g)} \in \mathbb{R}^3$, $N$ is the vertex number of $\mathcal{M}^{(g)}$ and similarly anchor points $\mathcal{A}^{(r)}$ for receiver's hand mesh. The object mesh $\mathcal{O}=\{O_k\}_{k=1}^M$, where $O_k \in \mathbb{R}^3$ and $M$ is the vertex number of the object mesh. The contact loss between the receiver's hand, the object, and the giver's hand is given by:
\begin{equation}
    \mathcal{L}_{con} = \frac{1}{|\mathcal{A}^{(r)}|}\sum_{A\in \mathcal{A}^{(r)}} (\min_k ||A, O_k||_2 - \min_j ||A, v_j^{(g)}||_2)
\end{equation}
$\mathcal{L}_{con}$ minimizes the minimal distance between the receiver's hand and object, but maximize the minimal distance between the two hands. We consider the anchor point as the contact points to reduce the computational time.

Besides, we adopt the refinement layer as in \cite{ganhand} and have two additional loss functions $\mathcal{L}_{arc}$ and $\mathcal{L}_{\gamma}$ to further minimize the distance between hand tips and object surface. For the formulation of $\mathcal{L}_{arc}$ and $\mathcal{L}_{\gamma}$, please refer to \cite{ganhand}.

\paragraph{Interpenetration Loss} If interpenetration occurs, there should exist a subset of the anchor points of the receiver's hand (denoted as $\mathcal{A}_{in}^{(r)} \in \mathcal{A}^{(r)}$) inside the object mesh $\mathcal{O}$ or the giver's hand mesh $\mathcal{M^{(g)}}$. Then, the interpenetration loss is given by:

\begin{equation}
    \mathcal{L}_{int} = \frac{1}{|\mathcal{A}_{in}^{(r)}|}\sum_{A\in \mathcal{A}_{in}^{(r)}} (\min_k||A, O_k|| + \min_j ||A, v_j^{(g)}||_2).
\end{equation}
$\mathcal{L}_{int}$ minimizes the min distance between the anchor point inside and the object surface or giver's hand mesh surface, so that the points inside will be driven outwards.

\paragraph{The Overall Loss} By taking all the loss function into consideration, we have the overall loss function for RGPNet.
\begin{equation}
    \mathcal{L} = \mathcal{L}_{gen} + \lambda_1\mathcal{L}_{gp} + \lambda_2 \mathcal{L}_{con} + \lambda_3\mathcal{L}_{arc} + \lambda_4\mathcal{L}_{\gamma} + \lambda_5{L}_{int}
\end{equation}


\section{Experiments \& Results}
In the main paper, we report the results on the H2O dataset. For the results on the H2O-Syn dataset, please refer to the supplementary materials.

\subsection{Implementation Details}
The hand-object pose estimation module is trained independently on the H2O pre-handover subset. Once the module is trained, we fix the parameters to regard the module as an off-the-shelf tool. We provide several baseline methods for human-object pose estimation, whose performance will be reported in Table \ref{tab:hor}. For details, please refer to the supplementary materials.

After that, the details of the GAN module are listed as follows: 
Input image patches are resized to $256\times 256$. Learning rate lr=0.0001, batch size bs=32, $\lambda_1=10$, $\lambda_2=100$, $\lambda_3=0.01$, $\lambda_4=20$, $\lambda_5=2000$. The optimizer is SGD with momentum = 0.9.
The Generator is updated 5 times before the Discriminator is optimized once. The GAN module is trained with cosine LR scheduling for 30 epochs. 
For more details on implementation, please refer to the supplementary materials.

\subsection{Metrics}
We adopt the following methods to measure the result qualities.\\
\textbf{Hand Pose Error} We compute the mean 3D vertex errors for hand to assess the quality of pose estimation.\\
\textbf{Object Pose Error} We also compute the mean 3D vertex errors for object to assess the quality of pose estimation.\\
\textbf{Grasp Score} We apply the widely-used analytical grasp metric from \cite{grasp_score} to measure the grasp quality. It indicates the minimum force to be applied to break the grasp stability.\\
\textbf{Interpenetration Volume} Similarly to \cite{cpf}, we voxelize the object mesh into $80^3$ voxels, and calculate the sum of the voxel volume inside the hand surface.

\subsection{Results on H2O Dataset}
\paragraph{Hand-object Interaction Reconstruction}
For the hand-object interaction reconstruction task, we adopt two baseline methods and since the training pipeline does not require information from the corresponding physical handover stage, and only the H2O pre-handover subset is required. The train/val/test split is kept the same as described in Sec. \ref{sec:statistic}. 
\begin{table}[]
    \centering
    \begin{tabular}{c|c|c}
        \hline
        Method &  Object error & Hand error\\ \hline
        Tekin \etal\cite{ho_recon1} & 28.4  & 19.2\\
        Hasson \etal\cite{ho_recon_photo} & 26.1 & 22.5\\\hline
    \end{tabular}
    \caption{Hand-object reconstruction results on H2O pre-handover subset, the errors are measure in mm. }
    \label{tab:hor}
\end{table}
To note, the baseline Hasson \etal listed is a variant without the photometric consistency loss \cite{ho_recon_photo}. Because we think the static prediction is more challenging, and many perception tasks are carried out under the single image setting.

\begin{figure*}[]
\centering
\includegraphics[width=\linewidth]{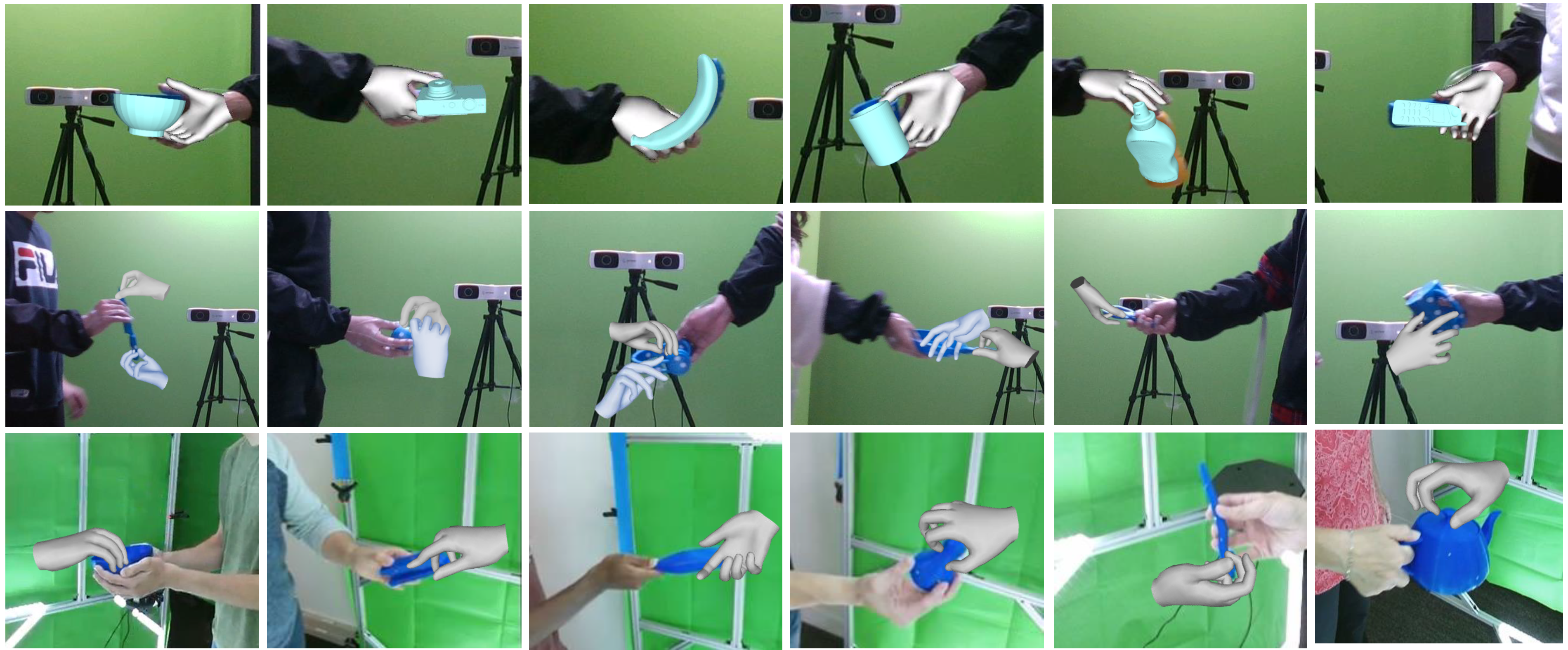}
   \caption{Qualitative results: \textbf{Upper:} Hand-objet interaction reconstruction results on our H2O pre-handover dataset; \textbf{Middle:} RGPNet results on H2O pre-handover dataset; \textbf{Bottom:} RGPNet results on ContactPose hand-off dataset. To note, for better visual quality, the displayed images are slightly larger than the image patch for training.}
\label{fig:qualitative}
\end{figure*}

Qualitative results on H2O dataset are shown in Fig. \ref{fig:qualitative}.

\paragraph{Receiver Grasp Prediction}
For the RGP task, we report the grasp score and interpenetration volume (denoted as "Interp.") in  Table \ref{tab:rgpr}. As there are no baselines for this task, we implement a variant of RGPNet which adopts the grasp type prediction. The RGPNet with grasp type prediction only add a grasp type branch after the global feature $\mathcal{F}$, and the predicted hand pose is $H=H_{pre} + H_o + \Delta H$, instead of the $H = H_{pre} + \Delta H$. $H_o=\{\beta_o, \theta_o, \bm{0}\}$ is the default hand pose for different grasp types. For the exact structure, please refer to the supplementary files.
\begin{table}[]
    \centering
    \begin{tabular}{c|c|c}
        \hline
       Method &  Grasp Score $\uparrow$& Interp. $\downarrow$\\ \hline
        RGPNet & 0.58 & 28\\
        RGPNet + grasp type & 0.62 & 24\\\hline
    \end{tabular}
    \caption{Receiver grasp prediction results. $\uparrow$ means the the higher the better; $\downarrow$ means the lower the better. }
    \label{tab:rgpr}
\end{table}

Besides, we apply the model trained on our H2O dataset to the ContactPose hand-off subset. The qualitative results are shown in Fig. \ref{fig:qualitative}.

\subsection{Transfer Human Demonstration to Robot Hand}

\begin{figure}[]
\centering
\includegraphics[width=\linewidth]{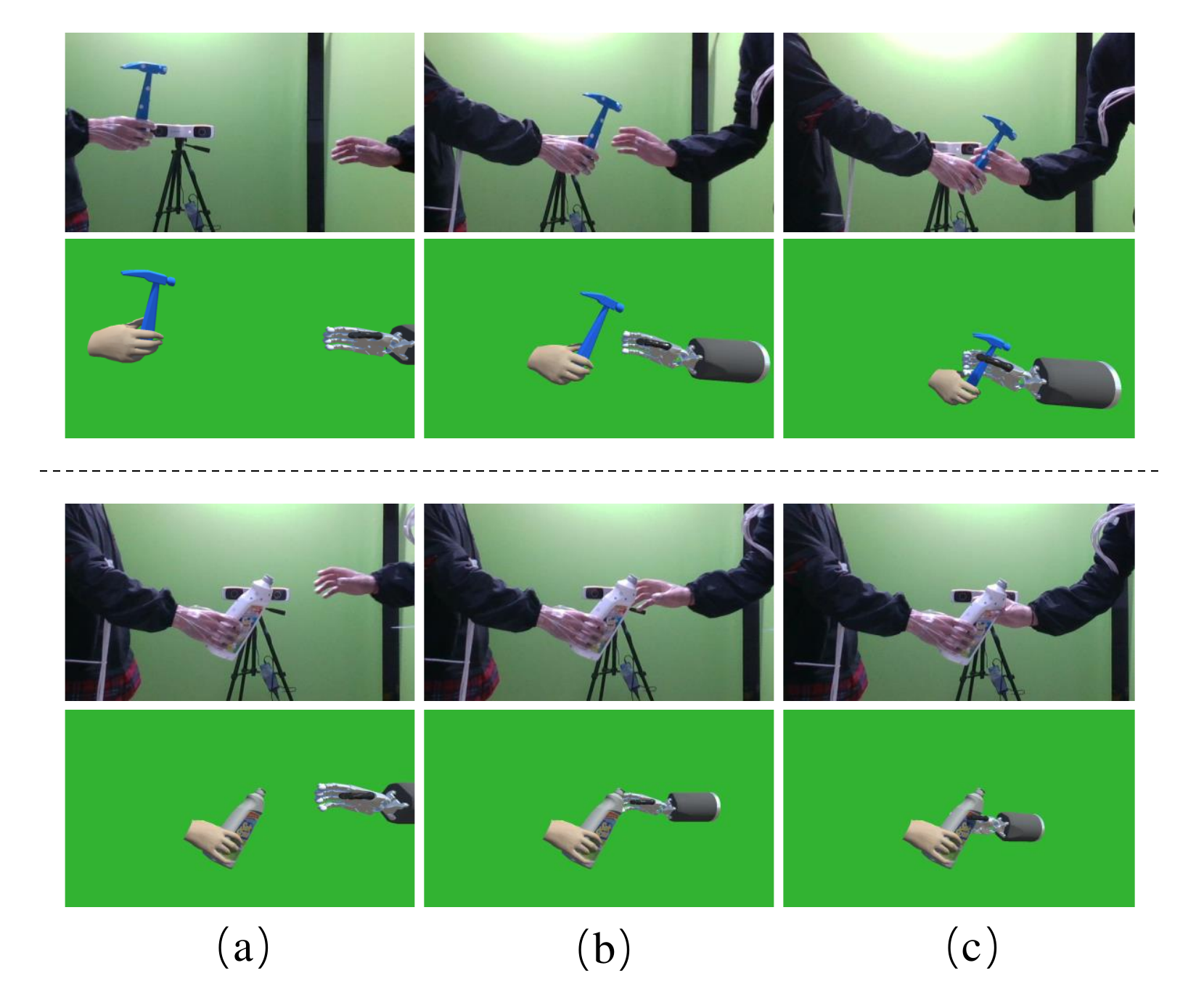}
   \caption{The H2O dataset can also be used as video demonstration for imitation learning. Above image content: (a)(b) are in the pre-handover phase and (c) is in the physical handover phase.}
\label{fig:imitation}
\end{figure}

Since human beings are generally the experts for robots to learn manipulation skills from, the handover video clips can be used as video demonstration sources. In this work, we make simple illustration on how to construct a human-robot handover demonstration by transferring the hand-object pose recordings to a simulator (Unity in the showcase).   To note, it is trivial to adopt the same process to construct a robot-human handover demonstration. Samples from two handover process including pre-handover and physical handover are displayed in Fig. \ref{fig:imitation}. The robot hand adopted for illustration is the Shadow dexterous hand\footnote{https://www.shadowrobot.com/dexterous-hand-series/}.
Conducting an imitation learning for real robot hand is beyond the scope of this paper, so we consider it as a future work.

\section{Conclusion and Future Works}
In this work, we propose a novel dataset with rich annotations for visual analysis of the handover tasks. To complete the visual analysis, we propose a novel RGPNet to address the seldom explored task Receiver Grasp Prediction. However, we need to emphasize that our H2O dataset is multi-purpose and can support various handover-related tasks, such as hand-object pose estimation, hand pose estimation. In the end, we showcase the potential use of our dataset as the video demonstration for the imitation learning of robot handover task or manipulation tasks. In the future, we would like to explore better visual perception algorithms for hand/object-related tasks. Besides, we are definitely eager to apply our datasets as the demonstration to help the robot learning.

{\small
\bibliographystyle{ieee_fullname}
\bibliography{hand}
}

\end{document}